\documentclass[sigconf]{acmart}

\usepackage{amsthm}
\usepackage{bbm}
\usepackage[linesnumbered,ruled]{algorithm2e}

\usepackage{amsmath,amsfonts,bm}

\def\eqref#1{equation~\ref{#1}}

\def\1{\bm{1}}

\def\mA{{\bm{A}}}

\def\mD{{\bm{D}}}

\def\mH{{\bm{H}}}

\def\mX{{\bm{X}}}

\def\mZ{{\bm{Z}}}

\DeclareMathAlphabet{\mathsfit}{\encodingdefault}{\sfdefault}{m}{sl}
\SetMathAlphabet{\mathsfit}{bold}{\encodingdefault}{\sfdefault}{bx}{n}

\AtBeginDocument{%
  \providecommand\BibTeX{{%
    \normalfont B\kern-0.5em{\scshape i\kern-0.25em b}\kern-0.8em\TeX}}}

\acmConference[FedGraph '22]{The First International Workshop on Federated Learning over Graph Data}{October 21, 2022}{Atlanta, GA, USA}
\acmBooktitle{The First International Workshop on Federated Learning over Graph Data (FedGraph '22), October 21, 2022, Atlanta, GA, USA}

\begin{document}

\title{Federated Graph Representation Learning \\using Self-Supervision}

\author{Susheel Suresh}
\authornote{Work performed during internship at Microsoft.}
\affiliation{%
  \institution{Purdue University}
  \country{USA}
}
\email{suresh43@purdue.edu}

\author{Danny Godbout}
\affiliation{
    \institution{Microsoft}
    \country{USA}
}
\email{danny.godbout@microsoft.com}

\author{Arko Mukherjee}
\affiliation{
    \institution{Microsoft}
    \country{USA}
}
\email{arko.mukherjee@microsoft.com}

\author{Mayank Shrivastava}
\affiliation{
    \institution{Microsoft}
    \country{USA}
}
\email{mayank.shrivastava@microsoft.com}

\author{Jennifer Neville}
\affiliation{%
  \institution{Purdue University / Microsoft}
  \country{USA}
}
\email{jenneville@microsoft.com}

\author{Pan Li}
\affiliation{%
  \institution{Purdue University}
  \country{USA}
}
\email{panli@purdue.edu}

\begin{abstract}
Federated graph representation learning (FedGRL) is an important research direction that brings the benefits of distributed training to graph structured data while simultaneously addressing some privacy and compliance concerns related to data curation. However, several interesting real-world graph data characteristics viz. label deficiency and downstream task heterogeneity are not taken into consideration in current FedGRL setups. In this paper, we consider a realistic and novel problem setting, wherein cross-silo clients have access to vast amounts of unlabeled data with limited or no labeled data and additionally have diverse downstream class label domains. We then propose a novel FedGRL formulation based on model interpolation where we aim to learn a shared global model that is optimized collaboratively using a self-supervised objective and gets downstream task supervision through local client models. We provide a specific instantiation of our general formulation using BGRL a SoTA self-supervised graph representation learning method and we empirically verify its effectiveness through realistic cross-slio datasets: (1) we adapt the Twitch Gamer Network which naturally simulates a cross-geo scenario and show that our formulation can provide consistent and avg. 6.1\% gains over traditional supervised federated learning objectives and on avg. 1.7\% gains compared to individual client specific self-supervised training and (2) we construct and introduce a new cross-silo dataset called Amazon Co-purchase Networks that have both the characteristics of the motivated problem setting. And, we witness on avg. 11.5\% gains over traditional supervised federated learning and on avg. 1.9\% gains over individually trained self-supervised models. Both experimental results point to the effectiveness of our proposed formulation. Finally, both our novel problem setting and dataset contributions provide new avenues for the research in FedGRL. 

\end{abstract}


\keywords{graph neural networks, federated learning, self-supervised learning}


\maketitle



\section{Introduction}
\label{sec:intro}
The widespread adoption of Graph neural networks (GNNs), a class of powerful encoders for graph representation learning~\cite{scarselli2008graph, kipf2017semi, hamilton2017inductive, chami2022machine} have shown enormous potential for downstream applications in a variety of domains spanning social, physical and biochemical sciences \cite{hamilton2020graph, GNNBook2022, senior2020improved, shlomi2020graph}. While GNNs have been extensively studied in both supervised \cite{dai2016discriminative, kipf2017semi, hamilton2017inductive, velivckovic2017graph, xu2018how, morris2019weisfeiler, li2020distance} and self-supervised \cite{velivckovic2018deep, hassani2020contrastive, you2020graph, suresh2021adversarial, xie2022self, liu2022graph} settings, the bulk of the work falls under a traditional data-centralized training regime. With heightened data security concerns, privacy, and compliance regulations in key GNN application domains such as social networks and healthcare, increasingly, vast amounts of such graph data is siloed away or held behind strict data boundary constraints~\cite{gdpr2019, custers2019eu}. Therefore, there is great need for understanding and developing decentralized training processes for GNNs.

Federated learning (FL)~\cite{mcmahan2017communication, li2020federated} has risen as a widely popular distributed learning approach that brings model training processes to the training data held at the clients, thereby avoiding transfer of raw client data. The benefits of FL are two fold, first, it is seen as a key ingredient in enabling privacy-preserving model learning in cross-geo or cross-silo scenarios~\cite{bonawitz2021federated}. Second, when certain participating clients have scarce training data or lack diverse distributions, FL enables them to potentiality leverage the power of data from others---thereby helping them improve performance on their own local tasks~\cite{mcmahan2017communication}. 

Recent research efforts have looked at applying federated learning algorithms to graph structured data~\cite{wang2020graphfl, he2021fedgraphnn, zhang2021subgraph, xie2021federated, liu2022federated, fu2022federated}. However, several interesting and real-world graph data characteristics are not taken into consideration: (1) \textbf{Label deficiency} - current methods assume that training labels (node / graph level) for the corresponding tasks~\footnote{e.g., classification / regression problems at node or whole graph level.} are readily available at the clients and a global model is trained end-to-end in a federated fashion. However, in many cross-silo applications, clients might have very little or no labeled data points. It is a well known that annotating labels of node / graph data takes a lot of time and resources~\cite{hu2019strategies, zitnik2018prioritizing}, e.g., difficulties in obtaining explicit user feedback in social network applications and costly \emph{in vitro} experiments for biological networks. Moreover, certain clients may be unwilling to share labels due to competition or other regulatory reasons. (2) \textbf{Downstream task heterogeneity} - it is reasonable to assume that while clients may share the same graph data domain, the downstream tasks may be client-dependent and vary significantly across clients. It is also reasonable to expect that some clients may have new downstream tasks added at a later point, where a model supervised by previous tasks may be ineffective.  

With these observations, we propose a realistic and unexplored problem setting for FedGRL: \emph{Participating clients have a shared space of graph-structured data, though the distributions may different across clients. And, clients have the access to vast amounts of unlabeled data. Additionally, they may have very different local downstream tasks with very few private labeled data points.} Fundamentally, our problem setting asks if one can leverage unlabeled data across clients to learn a shared graph representation (akin to ``knowledge transfer") which can then be further personalized to perform well in the local downstream tasks at each client.  In a data centralized training regime, a number of works that utilize GNN pre-training~\cite{hu2019strategies, qiu2020gcc} and self-supervision~\cite{thakoor2021bootstrapped, you2020graph, suresh2021adversarial, xie2022self} have shown the benefits of such approaches in dealing with label deficiency and transfer learning scenarios which motivate us to explore and utilize them for the proposed FedGRL problem setting. 

In this paper, we propose a novel FedGRL formulation based on model interpolation where we aim to learn a shared global model that is optimized collaboratively using a self-supervised objective and gets downstream task supervision through local client models. We provide a specific instantiation of our general formulation using BGRL~\cite{thakoor2021bootstrapped} a SoTA self-supervised graph representation learning method and we empirically verify its effectiveness through realistic cross-slio datasets: (1) we adapt the Twitch Gamer Network which naturally simulates a cross-geo scenario and show that our formulation can provide consistent and avg. 6.1\% gains over traditional supervised federated learning objectives and on avg. 1.7\% gains compared to individual client specific self-supervised training and (2) we construct and introduce a new cross-silo dataset called Amazon Co-purchase Networks that have both the characteristics of the motivated problem setting. We firstly show how standard supervised federated objectives can result in negative gains (on avg. -4.16\%) compared to individual client specific supervised training, due to the increased data heterogeneity and limited label availability. Then we experimentally verify the effectiveness of our method and witness on avg. 11.5\% gains over traditional supervised federated learning and on avg. 1.9\% gains over individually trained self-supervised models. Both experimental results point to the effectiveness of our proposed formulation. 

The remainder of this paper is organized as follows, in Sec.~\ref{sec:related} we review relevant work related to FL for graph structured data, self-supervised techniques for GNNs and finally some recent work on tackling label deficiency with FL. In Sec.~\ref{sec:prelim} we introduce notation and some preliminaries, later in Sec.~\ref{sec:method}, we provide a detailed problem setup, introduce our formulation and its instantiations. Later in Sec.~\ref{sec:exp_setup} we provide detailed experimental setup and finally present experimental results in Sec~\ref{sec:exp}.

\section{Related Work}
\label{sec:related}
The broad fields of designing GNNs for graph representation learning (GRL) get detailed coverage in recent surveys ~\cite{chami2022machine, hamilton2020graph, GNNBook2022}. We refer the reader to ~\cite{kairouz2021advances,li2020federated, li2020federatedopti} for a an overview of FL methods. 

\subsection{Federated Learning for Graphs}
FedGRL is a new research topic and current works have considered the following two main problem formulations.

First, for node-level tasks (predicting node labels), there are three sub categories based on the degree of overlap in graph nodes across clients: (1) No node overlap between client graphs. Here, each client maintains a GNN model which is trained on the local node labels and the server aggregates the parameters of the client GNN models and communicates it back in every federation round~\cite{zheng2021asfgnn, chen2021fedgraph, zhang2021subgraph}. ASFGNN~\cite{zheng2021asfgnn} additionally tackles the non-IID data issue using split based learning and FedGraph~\cite{chen2021fedgraph} focuses on efficiency and utilizes a privacy preserving cross-client GNN convolution operation. FedSage~\cite{zhang2021subgraph} considers a slightly different formulation, wherein each client has access to disjoint subgraphs of some global graph. They utilize GraphSage~\cite{hamilton2017inductive} and train it with label information and further propose to train a missing neighbor generator to deal with missing links across local subgraphs. (2) Partial node overlap across clients. Here, each participating client holds subgraphs which may have overlapping nodes with other clients graphs. GraphFL~\cite{wang2020graphfl}  considers this scenario and utilizes a meta-learning based federated learning algorithm to personalize client models to downstream tasks. \cite{peng2021differentially} considers overlapping nodes in local client knowledge graphs and utilize them to translate knowledge embedding across clients. (3) Complete node overlap across clients. Here all clients hold the same set of nodes, they upload node embeddings instead of model parameters to the server for FL aggregation. Existing works focus on the vertically partitioned citation network data ~\cite{zhou2020vertically, ni2021vertical}. Note that all the above problem settings are different from ours in motivation as we focus on label deficiency and downstream task heterogeneity.

Secondly, for graph-level tasks (predicting graph labels), each client has a local set of labeled graphs and the goal is to learn one global model or personalized local models using federation. This problem setting is fundamentally similar to other federated learning settings widely considered in vision and language domains. One needs to replace the previous linear/DNN encoder into a graph kernel/GNN encoder to handle the graph data modality. \cite{he2021fedgraphnn} creates a benchmark towards this end. The issues of client data non-IID ness carry over to the graph domain as well and \cite{xie2021federated} utilizes client clustering to aggregate model parameters.

\subsection{Self-Supervised Learning for GNNs}
Early SSL techniques for GNNs adopted the edge-reconstruction principle, where the edges of the input graph are expected to be reconstructed based on the output of GNNs~\cite{hamilton2017inductive,kipf2016variational}. Recently, contrastive methods~\cite{chen2020simple} effective on images have been successfully adapted to self-supervised GNN training. The main focus of these methods are in designing contrastive pairs for irregular graph structures unlike images and thus becomes more challenging. Some works use different parts of a graph to build contrastive pairs, including nodes v.s. whole graphs~\cite{sun2019infograph, velivckovic2018deep}, nodes v.s. nodes ~\cite{peng2020graph}, nodes v.s. subgraphs~\cite{jiao2020sub}. Other works have also employed graph data augmentation like edge perturbation, node dropping, subgraph sampling for building contrastive pairs and have been commonly applied for both node and graph level representation learning~\cite{you2020graph,hassani2020contrastive, suresh2021adversarial, zhu2020deep}. However, due to the need for costly negative sampling to generate contrastive pairs, large negative examples and batch sizes, these contrastive based methods suffer from scalability and efficiency issues when applied to large scale graphs~\cite{thakoor2021bootstrapped}. Recent efforts have adapted Siamese representation learning techniques~\cite{chen2021exploring} that do not rely on negative examples, but contrast representations from GNN encoders on different augmentations of a graph for e.g., BGRL~\cite{thakoor2021bootstrapped}, SelfGNN~\cite{kefato2021self} and DGB~\cite{che2021self}. It is important to note that our general self-supervised FedGRL formulation can be instantiated using any of the above SSL techniques developed for GNNs.

\subsection{Label Deficiency and FL}
There are also some very recent developments in handling the label deficiency issue within FL mainly for vision domains that are relevant to our work~\cite{zhang2020federated, zhuang2021collaborative, zhuang2022divergence, he2021ssfl, lubana2022orchestra}. FedCA~\cite{zhang2020federated} and FedU~\cite{zhuang2021collaborative} are based on direct extension of self-supervised methods SimCLR~\cite{chen2020simple} and BYOL~\cite{grill2020bootstrap} to the federated regime respectively. FedEMA~\cite{zhuang2022divergence} extends FedU to further address the non-IID data issue by controlling the divergence between global and local models dynamically using exponential moving average. Orchestra~\cite{lubana2022orchestra} also employs an SSL based method during federation and addresses the non-IID problem using client clustering. 
SSFL~\cite{he2021ssfl} utilizes SimSiam~\cite{chen2021exploring}, a siamese SSL method in conjunction to both standard and personalized FL algorithms viz., FedAvg~\cite{mcmahan2017communication}, perFedAvg~\cite{fallah2020personalized}, Ditto~\cite{li2021ditto}. While these methods have utilized and shown the benefits of using unlabeled data within federated learning for image domain, they have not shown the experimental efficacy of their methods for real world datasets with downstream task heterogeneity across clients, which heterogeneity graph learning tasks often contain.

\section{Preliminaries}
\label{sec:prelim}
We consider a graph $G =(V, E)$ with node set $V = \{v_1, \dots v_N\}$ and edge set $E \subseteq V \times V$. Additionally, we use node feature matrix $\mX \in \mathbb{R}^{N \times F}$ and adjacency matrix $\mA \in \{0,1\}^{N \times N}$ where, $N = |V|$ is the number of nodes in the graph, $F$ is the node feature dimension and $\mA_{i,j} = 1$ iff $(v_i, v_j) \in E$.

\subsection{Graph Neural Networks}
A GNN encoder $f(\mX, \mA)$ or equivalently\footnote{later we omit $\mX$ and $\mA$ and simply denote the input as a graph G for brevity.} $f(G)$, receives the node features and graph structure to produce node representations $\mH \in \mathbb{R}^{N \times F^\prime}$ of dimension $F^\prime \ll F$ which is defined using a layer-wise propagation rule for e.g., GCN~\cite{kipf2017semi} defined as follows: 
\begin{equation}
    \mH_{(l+1)} = \sigma \Big( \tilde{\mD}^{-\frac{1}{2}} \tilde{\mA} \tilde{\mD}^{-\frac{1}{2}} \mH_{(l)} W_{(l)}\Big)
\end{equation}
where, $\tilde{\mA} = \mA + I$ is the adjacency matrix with added self loops, $\tilde{\mD}$ is the degree matrix with $\tilde{\mD}_{ii} = \sum_{j} \tilde{\mA}_{ij}$. $W_{(l)}$ is trainable weight matrix for layer $l$ and $\mH_{(l)}$ is the intermediate $l^{\text{th}}$ layer node representation matrix with $\mH_{(0)} = \mX$. We omit the layer index and use $\mH = \mH_{(l+1)}$ to denote the final node representations produced by the $l$-layer GNN. These representations can be used for downstream tasks, such as node classification. 

\subsection{Federated Optimization}

In standard federated optimization, we are given a set of clients $C$. A client $c \in C$ holds $n_c$ amount of private data drawn from distribution $\mathcal{D}_{c}$ and the goal is to learn $f_c : \mathcal{X}_c \rightarrow \mathcal{Y}_c$ for each $c$ and the overall distributed optimization is of the form~\cite{mcmahan2017communication}: 
\begin{equation}\label{eq:fed_1}
    \min_{\{f_c\}} \sum_{c \in \mathcal{C}} \frac{n_c}{n} \mathcal{L}_c(f_c)
\end{equation}
where, $n = \cup_{c \in \mathcal{C}} n_c$ is total amount of data. For each client c, the expected loss over its data distribution is,
\begin{equation}\label{eq:fed_2}
    \mathcal{L}_c (f_c) = \mathbb{E}_{(x_c, y_c) \sim \mathcal{D}_c} [\ell_{c} (f_c, x_c, y_c)],
\end{equation}
where, $(x_c, y_c)$ represents data and labels in client c and $\ell_{c} (f, x_c, y_c)$ is the loss function for e.g., cross-entropy loss for classification problems.

\section{Method}
\label{sec:method}
In this section, we describe the problem setting for self-supervised FedGRL, provide a general framework and then introduce a specific instantiation. 

\subsection{Problem Setting}
\begin{figure}[h]
    \centering
    \includegraphics[width=0.45\textwidth]{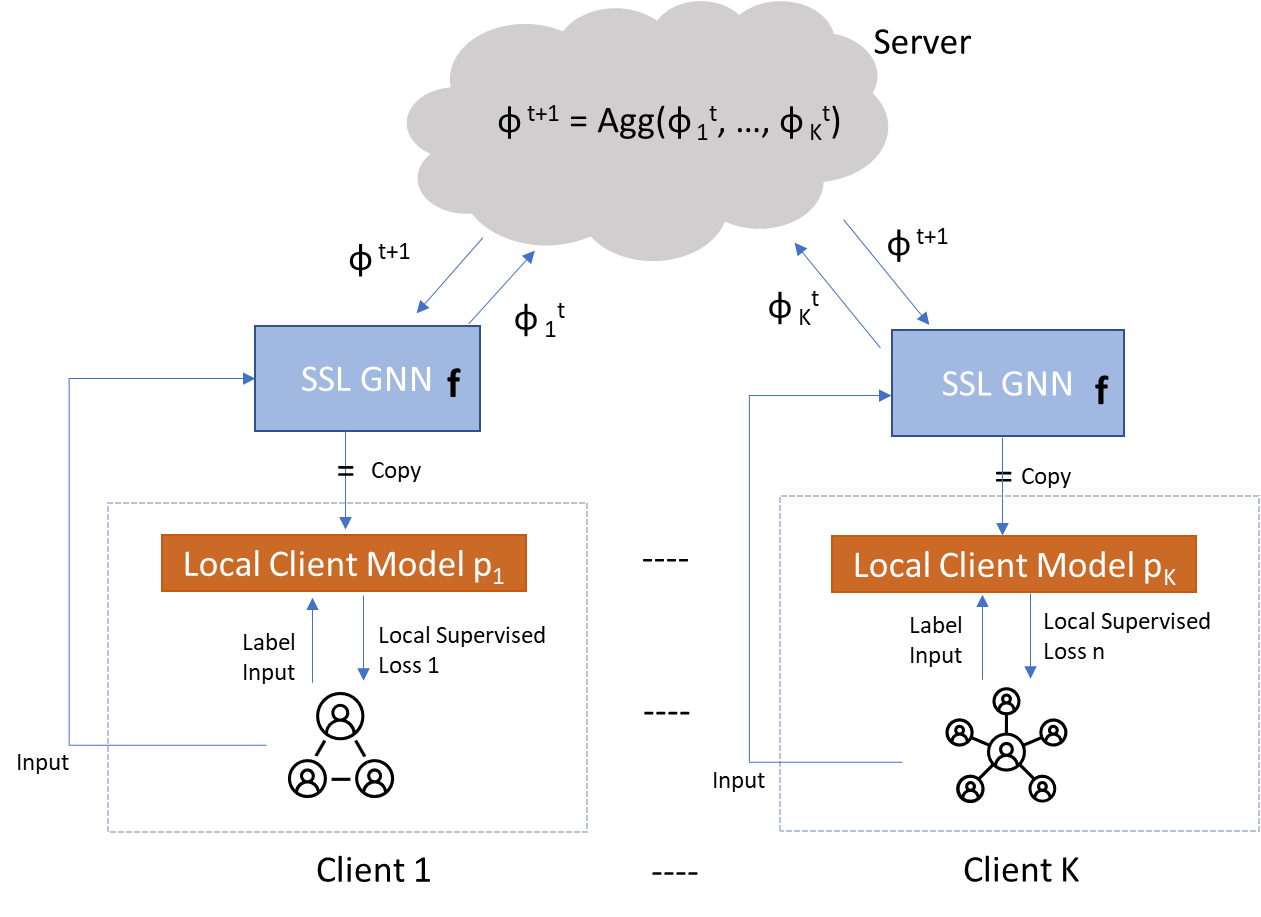}
    \caption{Overview of the proposed general formulation for self-supervised FedGRL. Here, we learn a shared global GNN model $f$ using a self-supervised (SSL) objective, collaboratively based on federated learning. With the copy gate, $f$ is only trained on unlabeled data through SSL and local client models $p_c$ are further trained individually on top of $f$ with label supervision. This forms our first learning protocol (\textbf{LP1}) . Without the copy gate, in addition to SSL objective, $f$ can further receive label supervision from client tasks and this forms the second learning protocol (LP2).}
    \label{fig:formulation}
\end{figure}

Each client $c \in \mathcal{C}$ holds a graph $G_c = (V_c, E_c)$ and has a downstream task $T_c$ if one exists. Let $N_c = |V_c|$ be the number of nodes in client graph. The node set $V_c$ is partitioned into labeled and unlabeled nodes as, $V_c = {V_c^{l} \cup V_c^{u}}$. If a certain client has no downstream task, then $V_c^{l} = \emptyset$. A client c with task $T_c$ has access to a space of node label $\mathcal{Y}_c$ of size $m_c$. Different clients may have entirely different tasks and thus $\mathcal{Y}_c$'s may not overlap across clients. Each node in $V_c^l$ gets a label from $\mathcal{Y}_c$ to form its corresponding set of node labels $Y_c$. The task $T_c$ is then to infer the labels for unlabeled nodes in $V_c^u$. Note that there could be multiple tasks for each client but it will not
change the foundation of the problem. So, we simply consider one task per client.

There are several differences compared to the traditional FL setting: (1) Label scarcity i.e., for each client c, $|V_c^l|$ could be very small or even zero. (2) Total labeled data may be far less than unlabeled data i.e., $\sum_{c} |V_c^l| \ll \sum_{c} |V_c^u|$ (3) Poor data quality i.e., for some clients $|V_c^l| + |V_c^u|$ may also be small. (4) Downstream task heterogeneity i.e., across clients, the class label domain $\mathcal{Y}_c$'s could be very different.

\subsection{General Formulation}
Directly optimizing $\{f_c\}_{c \in C}$ using the standard FL optimization in Eq.~\ref{eq:fed_1} and \ref{eq:fed_2} with labels, may not achieve good results as $\sum_{c} |V_c^l| \ll \sum_{c} |V_c^u|$. So, instead, we propose to utilize a model interpolation based formulation where a global shared model and client specific local models are simultaneously optimized using self-supervision and label supervision if downstream task labels are available. Specifically, we set $f_c = p_c \circ f$, where $p_c$ is the local client model, $f$ is the global shared model. We then propose to empirically solve the following distributed objective:
\begin{equation}\label{eq:OBJ}
    \min_{\{p_c\}, f} \sum_{c \in \mathcal{C}} \left[ \mathcal{L}_c (p_c, f, G_c, Y_c) \mathbbm{1}_{T_c\;\text{exists}}  + \lambda_c \tilde{\mathcal{L}_c} ( f, G_c)\right]
\end{equation}
where, the first term is the objective supervised by labels and the second term is the self-supervised objective that doesn't utilize label information, $\lambda_c$ controls the amount of model interpolation. Note that $f$ is shared while $p_c$ is not shared. It is important to point out that, by setting each $\lambda_c$ to 0.0, the objective in Eq.~\ref{eq:OBJ} falls back to standard supervised FL optimization. An overview of the formulation is shown in Fig~\ref{fig:formulation}. From the general formulation in Eq.~\ref{eq:OBJ}, we propose two practical learning protocols (LPs). 

\begin{itemize}
    \item \textbf{LP1}: The first supervised objective term is ignored, each $\lambda_c$ is set to 1.0 and a global model $f$ is learnt in a federated self-supervised fashion using the objective in the second term. Notice that here, labeled data is not used to train $f$ during federation. Once such a $f$ is trained, each client $c$ can further train a task specific local model $p_c$ themselves again by \textbf{freezing} or \textbf{finetuning} on top of $f$.
    \item \textbf{LP2}: Both terms are used with a non zero $\lambda_c$ during federation. Both $p_c$ and $f$ get label data supervision if available through the first term. Additionally, $f$ also gets trained using the self-supervised objective through the second term.
\end{itemize}

We will \textbf{only} consider the first learning protocol \textbf{(LP1)} for further experimentation in Sec~\ref{sec:exp} and leave the study of \textbf{LP2} for future work.

\subsection{Self-Supervised Objective}
\begin{figure}
    \centering
    \includegraphics[width=0.40\textwidth]{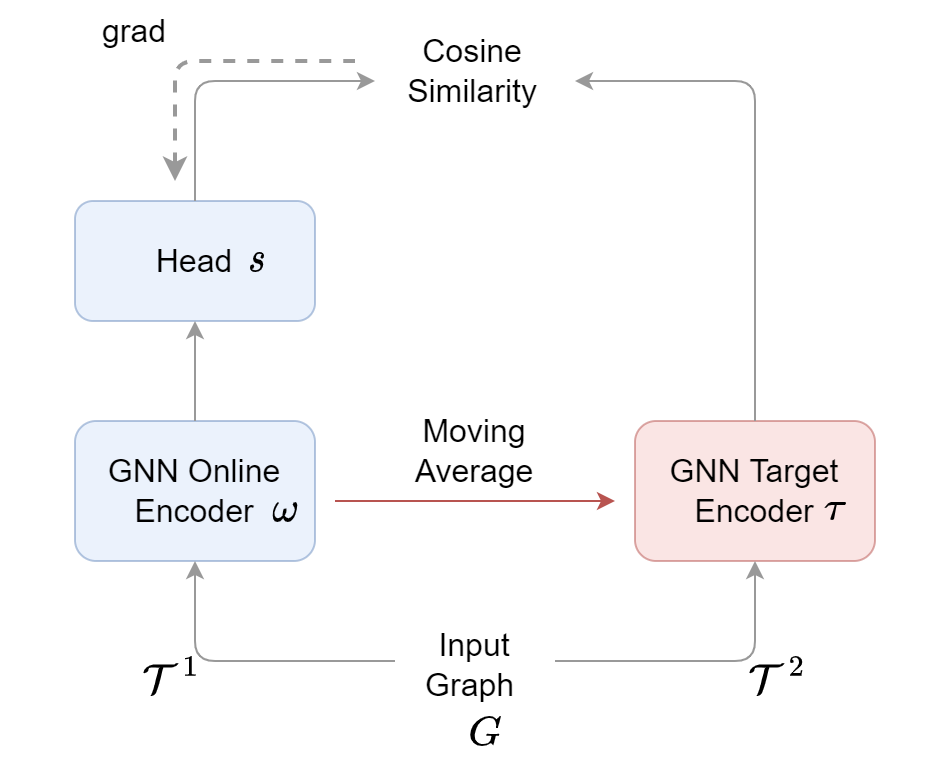}
    \caption{Overview of BGRL~\cite{thakoor2021bootstrapped}. Here, two correlated views of input $G$ are obtained using augmentations $\mathcal{T}^1$ and $\mathcal{T}^2$. Further, two GNN encoders $\omega$ and $\tau$ are used to obtain node representations. The node level head $s$ is tasked at using the representations from $\omega$ to predict the representations obtained from $\tau$ using a cosine similarity metric. Gradients flow only through $s$ and $\omega$ and $\tau$'s parameters are updated using an exponential moving average of $\omega$'s parameters. }
    \label{fig:bgrl}
\end{figure}

While in theory, the self-supervised loss term $\tilde{\mathcal{L}_c} (\cdot)$ in Eq~\ref{eq:OBJ} can be instantiated by using any self-supervised objectives, for practical reasons, more considerations are needed. Methods like GRACE~\cite{zhu2020deep} and GCA~\cite{zhu2021graph} learn node representations by maximizing the agreement between the same node pairs from two augmented versions of a graph while simultaneously minimizing the agreement between every other node pair. This leads to a scalability issue on large graphs as they require costly (time and space complexity quadratic in the number of nodes) negative sampling for building contrastive pairs~\cite{thakoor2021bootstrapped}. Moreover, in practice these methods rely on a large number of such negatives to be effective. We adopt BGRL~\cite{thakoor2021bootstrapped} which is a scalable and SoTA for node level self-supervised learning of GNNs. BGRL scales linearly with the the number of nodes in the input graph as it does not require contrasting negative node pairs. Fig.~\ref{fig:bgrl} shows an overview of the approach. 

\subsubsection{Instantiating using BGRL}
We describe the method for a single client and omit the client subscript $c$ for brevity. Specifically, BGRL first produces two alternate but correlated views of the input graph $G$: $G^1 = \mathcal{T}^1 (G)$ and $G^2 = \mathcal{T}^2 (G)$, by using stochastic graph augmentations $\mathcal{T}^1$ and $\mathcal{T}^2$ respectively. Examples include node feature masking and edge masking. BGRL, then utilizes two GNN encoders, an online $\omega(\cdot)$ and a target $\tau(\cdot)$ encoder. The online encoder produces online node representations for the first graph  $\mH^1 = \omega(G^1)$ and similarly, the target encoder produces target node representations of the second augmented graph as, $\mH^2 = \tau(G^2)$. Then, the online node representations are fed to a node level head $s(\cdot)$ that outputs predictions of the target node representations as $\mZ^1 = s(\mH^1)$. Then, the self-supervised BGRL objective is as follows,

\begin{equation}
    \tilde{\mathcal{L}} (\omega, s, \tau) = -\frac{2}{N} \sum_{i=0}^{N-1} \frac{\mZ^1_i \; {\mH^2_i}^{\top} }{\Vert \mZ^1_i \Vert \Vert \mH^2_i\Vert}
\end{equation}
where $N$ is the number of nodes in $G$. It is important to note that gradients of the above objective are taken only w.r.t the parameters of $\omega(\cdot)$ and $s(\cdot)$ and only they get optimized by gradient descent. The parameters of $\tau(\cdot)$ are updated using an exponential moving average of the parameters of $\omega(\cdot)$~\cite{thakoor2021bootstrapped}. 

During federation, parameters of online GNN encoder $\omega(\cdot)$ and node level head $s(\cdot)$ are both shared across clients and the server aggregates them in each federation round and sends it back to the clients. We refer to the global shared model $f = \omega$ and is used for the supervised objectives and further downstream tasks. The target GNN encoder $\tau(\cdot)$ is an auxiliary unit and is only used locally within the client for the purpose of  self-supervised learning and discarded once training is finished. 

\subsection{Task Specific Supervision}
The first loss term $\mathcal{L}_c (\cdot)$ utilizes supervision from the node labels if available. Standard cross-entropy loss over all labeled nodes can be utilized as commonly employed in semi-supervised GNN node classification~\cite{kipf2017semi}. For completeness it takes the following form,

\begin{equation}
    \mathcal{L} (p_c, f, G_c, Y_c) = - \sum_{i \in V_c^{l}} \sum_{j = 1}^{m_c} {Y_c}_{(ij)} \ln {\mZ_c}_{(ij)}
\end{equation}

where, $\mZ_c = \text{softmax}(p_c \circ f (G_c))$ and $\mZ_c \in \mathbb{R}^{N_c \times m_c}$.

\section{Experimental Setup}
\label{sec:exp_setup}
In this section, we first introduce the real-world datasets apt for our proposed problem setting. We then describe the baseline methods, experiment protocol, model architecture and hyper-parameters. 

\subsection{Datasets}
\label{sec:datasets}

\subsubsection{Twitch Gamer Networks}
These are a set of six user-user social network of gamers who stream in certain countries~\cite{musae}. Nodes are users and edges represent friendships. Node features are 128-dimensional vectors based on games liked and played, streaming habits and location. These networks are naturally formed in different geographies and share the same feature space which allows us consider them as a cross-silo scenario for our problem setting of federated "knowledge transfer". Additionally, each network also has a binary node classification task of predicting if a user uses explicit language. This enables us to empirically evaluate the learnt representations. The statistics of these networks are provided in Table~\ref{tab:twitch_dataset}.

\begin{table*}[]
\centering
\caption{Statistics of Twitch Gamer Networks. Each network from a geography represents a client graph in our experiments. }
\label{tab:twitch_dataset}
\resizebox{0.7\textwidth}{!}{%
\begin{tabular}{@{}lcccccc@{}}
\toprule
\multicolumn{1}{c}{} & twitch-DE & twitch-EN & twitch-ES & twitch-FR & twitch-PT & twitch-RU \\ \midrule
\#nodes              & 9,498     & 7,126     & 4,648     & 6,549     & 1,912     & 4,385     \\
\#edges              & 153,138   & 35,324    & 59,382    & 112,666   & 31,299    & 37,304    \\
density              & 0.003     & 0.002     & 0.006     & 0.005     & 0.017     & 0.004     \\ \bottomrule
\end{tabular}%
}
\end{table*}

\subsubsection{Amazon Co-purchase Networks}
We construct a set of six co-purchase networks from  raw Amazon reviews\footnote{\url{https://nijianmo.github.io/amazon/index.html}} and product meta data~\cite{ni2019justifying}. Specifically, we first consider six top level product segments viz. computer, photo, phone, tool, guitar and art. For each segment we construct a co-purchase network where nodes represent various products in the chosen segment and edges signify a frequent co-purchase (using the "also buy" signal in the raw product meta data). Each of these networks is held privately by six different clients simulating a cross-silo scenario. We then build a common review vocabulary of all products considered across all the six networks and use 1000-dimensional bag-of-words product review encoding as node features. This ensures all nodes across graphs share the same node feature space to make federated "knowledge transfer" meaningful and apt to our problem setting. The downstream task for each network is to classify the products (nodes) to fine-grained sub-segments. Again, this construction naturally simulates downstream task heterogeneity as each graph has a different class label domain. The statistics of these networks are provided in Table~\ref{tab:amazon_dataset}.

\begin{table*}[]
\centering
\caption{Statistics of Amazon Co-purchase Networks. Refer to Fig.~\ref{fig:amazon_labels} for detailed label distributions.}
\label{tab:amazon_dataset}
\resizebox{0.6\textwidth}{!}{%
\begin{tabular}{@{}lcccccc@{}}
\toprule
          & computer & photo  & phone   & tool   & guitar & art    \\ \midrule
\#nodes   & 10,055   & 4,705  & 16,683  & 4,827  & 2,506  & 6,610  \\
\#edges   & 87,512   & 28,818 & 113,760 & 43,458 & 10,342 & 90,678 \\
\#classes & 9        & 8      & 7       & 6      & 5      & 8      \\ \bottomrule
\end{tabular}%
}
\end{table*}

\begin{figure}
    \centering
    \includegraphics[width=0.5\textwidth]{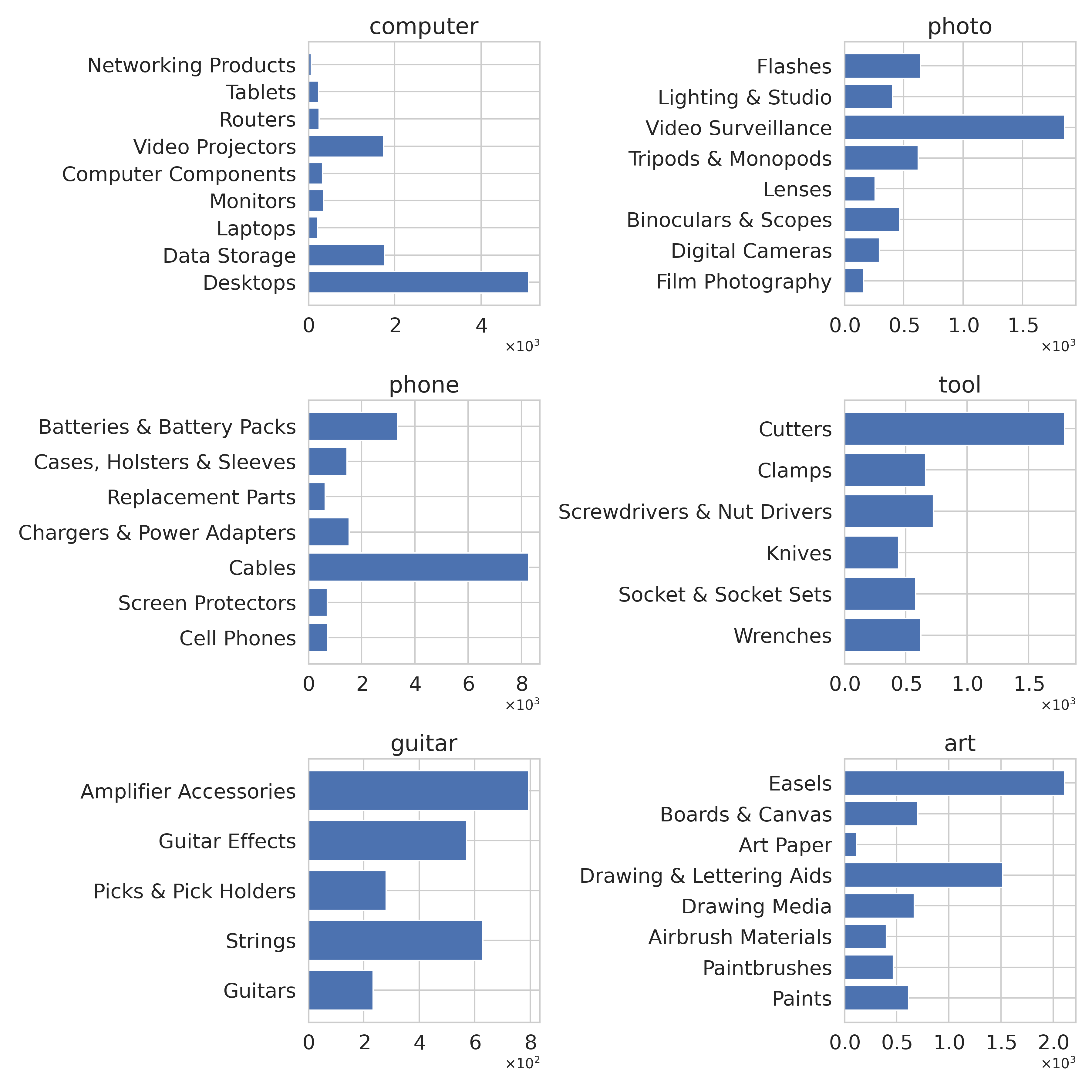}
    \caption{Class label distribution for Amazon Co-purchase Networks}
    \label{fig:amazon_labels}
\end{figure}

\subsection{Baselines}
For baselines, we consider: (1) \textbf{No-Fed-Sup} which trains a GNN model end-to-end with label supervision, for each client individually without federation. (2) \textbf{No-Fed-Self-Freeze} which first trains a GNN model using self-supervision and then performs linear evaluation after freezing the GNN node representations, for each client individually without federation. (3)  \textbf{No-Fed-Self-Finetune} which first trains a GNN model using self-supervision and then performs end-to-end task specific fine-tuning with a MLP task head on top, for each client individually without federation. (4) \textbf{Fed-Sup} trains a local GNN encoder + local MLP task head with end-to-end label supervision using federated learning. Only the parameters of the local GNN models are shared across clients and their parameters are aggregated in the server via  FedAvg~\cite{mcmahan2017communication} and sent back after every communication round.
We choose the \textbf{No-Fed} class of baselines to experiment whether FL can bring improvements to
each client through collaborative training. The \textbf{Freeze} and \textbf{Finetune} baselines allow us to experiment with the best downstream task evaluation strategy. The \textbf{Fed-Sup} baseline is the traditional FL technique that utilizes label information across clients. This baseline allows us to experiment if the formulation (and method) we propose to utilize self-supervision signal within FL brings improvements to each client. 

\subsection{Experiment Protocol}
For twitch network experiments, we use a 60/20/20 training, validation and test node random split for evaluation and for amazon network experiments, we use a 10/10/80 random split. Both experiments utilize F1-Micro score as the metric for node classification performance and we report average score results over 10 data splits.

\subsection{Parameter Settings}
For all baselines and our methods, we employ a 2-layer GCN~\cite{kipf2017semi} encoder with hidden dimensions tuned from $\{64, 128, 256\}$ and batch normalization. For supervised baselines \textbf{No-Fed-Sup} and \textbf{Fed-Sup} we use the AdamW~\cite{kingma2014adam} optimizer with learning rate of 0.01 and weight decay of 1e-3. For \textbf{No-Fed-Self} baselines and our \textbf{Fed-Self} methods we use the AdamW optimizer with a learning rate of 1e-4 annealed using a cosine scheduler as prescribed in ~\cite{thakoor2021bootstrapped} and weight decay of 1e-5. Additionally, the hidden dimension of the MLP predictor model $s(\cdot)$ in BGRL is set as 512. The exponential moving average decay rate in BGRL is initialized as 0.99 and gradually increased to 1.0 using cosine schedule. We perform a small random grid search (20 max trials) over node feature and edge masking probabilities from $\{0.1, 0.2, \dots, 0.8, 0.9\}$ for both augmentations $\mathcal{T}^1$ and $\mathcal{T}^2$. For all \textbf{Fed} methods the number of local epochs is set to be 1 and the number of communication rounds for \textbf{Fed-Sup} is set to 500 and \textbf{Fed-Self} is set to 10,000 with 1,000 warmup rounds. For \textbf{Freeze} linear evaluation, we employ a $\ell_2-$regularized logistic regression classifier and perform regularization strength search over $\{2^{-10}, 2^{-8}, 2^{-6}, \dots, 2^{6}, 2^{8}, 2^{10}\}$. For \textbf{Finetune} evaluation, we employ a task specific MLP head of 128 dimensions and perform optimization using AdamW for 100 steps with a learning rate of 0.01 and weight decay of 1e-3.

\section{Experimental Analysis}
\label{sec:exp}
In this section, we experimentally compare our instantiated methods against baselines for twitch and amazon networks.
\subsection{Twitch Network Experiments}

\begin{table*}[h]
\centering
\caption{Node classification task on Twitch Gamer Networks. Avg F1-Micro Scores over 10 splits and std. dev.}
\label{tab:twitch_results}
\resizebox{\textwidth}{!}{%
\begin{tabular}{lcccccc}
\hline
                         & twitch-DE         & twitch-EN         & twitch-ES         & twitch-FR         & twitch-PT         & twitch-RU         \\ \hline
No-Fed-Rand-Init-GNN     & 0.502 $\pm$ 0.103 & 0.501 $\pm$ 0.047 & 0.496 $\pm$ 0.198 & 0.510 $\pm$ 0.131 & 0.533 $\pm$ 0.156 & 0.507 $\pm$ 0.248 \\
No-Fed-Sup               & 0.673 $\pm$ 0.010 & 0.584 $\pm$ 0.014 & 0.725 $\pm$ 0.014 & 0.626 $\pm$ 0.016 & 0.678 $\pm$ 0.023 & 0.751 $\pm$ 0.015 \\
No-Fed-Self-Freeze       & 0.685 $\pm$ 0.009 & 0.617 $\pm$ 0.012 & 0.731 $\pm$ 0.010 & 0.626 $\pm$ 0.015 & 0.696 $\pm$ 0.021 & 0.752 $\pm$ 0.009 \\
No-Fed-Self-Finetune     & 0.703 $\pm$ 0.012 & \textbf{0.665 $\pm$ 0.023} & 0.746 $\pm$ 0.015 & 0.635 $\pm$ 0.018 & 0.710 $\pm$ 0.026 & \textbf{0.762 $\pm$ 0.011} \\ \hline
Fed-Sup                  & 0.677 $\pm$ 0.009 & 0.600 $\pm$ 0.009 & 0.723 $\pm$ 0.013 & 0.627 $\pm$ 0.018 & 0.683 $\pm$ 0.013 & 0.743 $\pm$ 0.015 \\
Fed-Self-Freeze (ours)   & 0.686 $\pm$ 0.007 & 0.606 $\pm$ 0.012 & 0.733 $\pm$ 0.007 & 0.626 $\pm$ 0.014 & 0.699 $\pm$ 0.022 & 0.752 $\pm$ 0.009 \\
\textbf{Fed-Self-Finetune} (ours) & \textbf{0.706 $\pm$ 0.013} &  0.657 $\pm$ 0.024 & \textbf{0.745 $\pm$ 0.013} & \textbf{0.636 $\pm$ 0.021} & \textbf{0.712 $\pm$ 0.024} & 0.761 $\pm$ 0.014 \\ \hline
\end{tabular}%
}
\end{table*}

\begin{figure}
    \centering
    \includegraphics[width=0.5\textwidth]{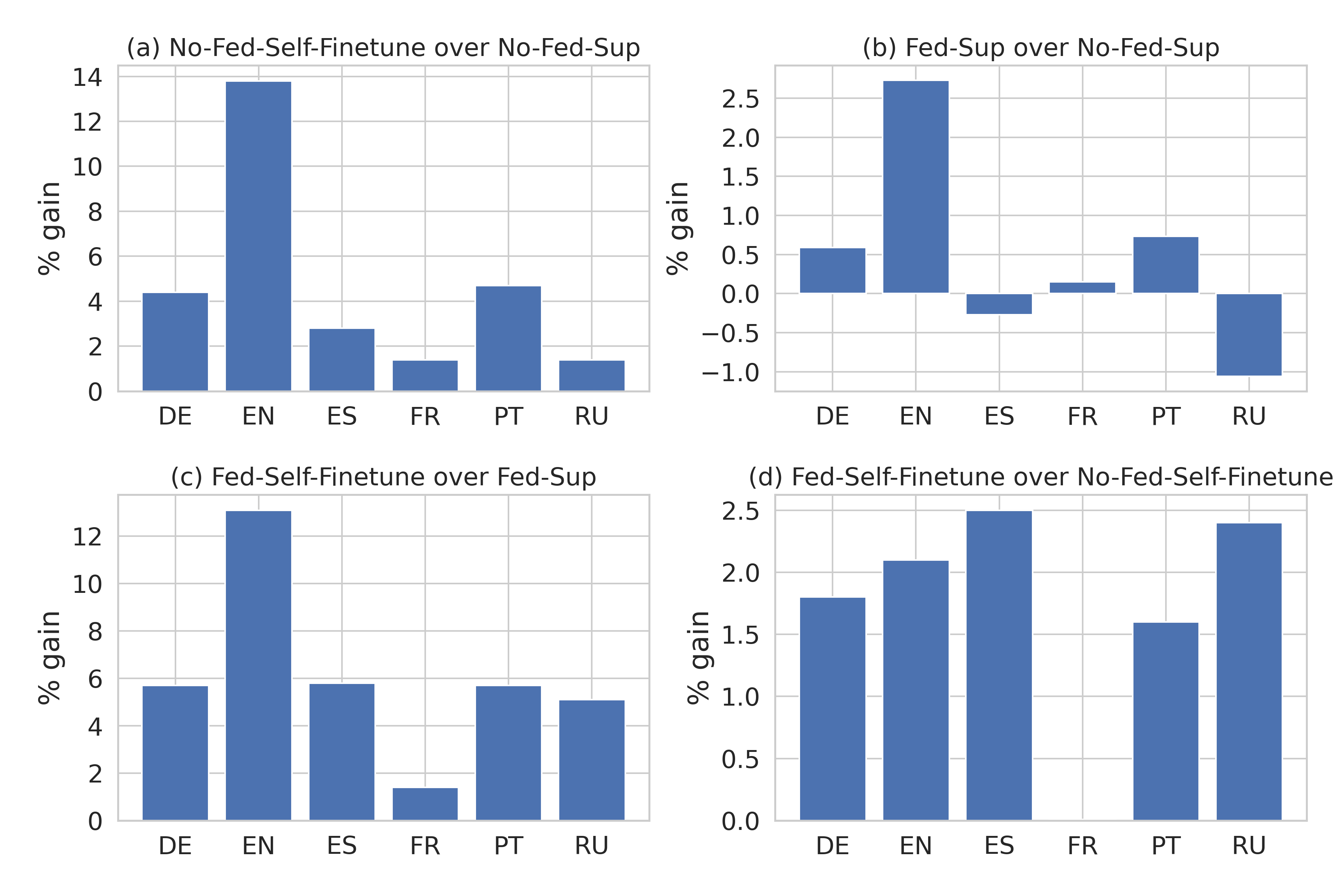}
    \caption{Performance gains in Twitch network experiments}
    \label{fig:gains_twitch}
\end{figure}

Table~\ref{tab:twitch_results} shows the results across all Twitch clients graphs. Firstly, it is clear that both No-Fed-Self-Freeze and No-Fed-Self-Finetune perform better than the No-Fed-Sup baseline for all the clients. This shows the usefulness of the self-supervised objective as it can make use of the available unlabeled data in each client albeit individually. Secondly, we witness that fine-tuning based evaluation (No-Fed-Self-Finetune) leads to the best gain of 4.75\% averaged across all clients over No-Fed-Sup. The individual client gains are shown Fig.~\ref{fig:gains_twitch}(a). Thirdly, Fed-Sup is only marginally better (0.45 \% avg. gain) than No-Fed-Sup, indicating that a supervised federated baseline is not effective in of properly utilizing all the client label information. In fact, for certain clients we see negative gains as shown in Fig~\ref{fig:gains_twitch}(b). Fourthly, our method \textbf{Fed-Self-Finetune} provides a gain of on avg. 6.13\% across all clients over Fed-Sup and Fig~\ref{fig:gains_twitch}(c) shows individual client gains we witness. This provides the evidence of the benefits in using self-supervised objective for learning the global model during federation. We reason that the heterogeneity across these client graphs is due to the fact that they are naturally formed in different geographies with diverse friendship cultures and language-use patterns. In such a scenario, standard supervised FedAvg suffers (Fed-Sup) due to the reliance on labels alone to learn representations, whereas utilizing unlabeled data and performing self-supervised FedAvg can learn shared patterns and provide more robust representations for each client (Fed-Self-Finetune). Finally, our method \textbf{Fed-Self-Finetune} provides consistent gains over individually trained No-Fed-Self-Finetune baseline and overall we witness an avg. 1.73\% gain across all clients (see also Fig~\ref{fig:gains_twitch}(d)). This further shows that we can indeed utilize unlabeled data collaboratively using federation and justifies for formulation. This can also be attributed statistically, as there is an increase in effective unlabeled data which the shared GNN encoder model can leverage.

\subsection{Amazon Network Experiments}

\begin{table*}[h]
\centering
\caption{Node classification task results on Amazon Co-purchase Networks. Avg. F1-Micro Scores over 10 splits and std. dev.}
\label{tab:amzn_results}
\resizebox{\textwidth}{!}{%
\begin{tabular}{lcccccc}
\hline
                         & computer          & photo             & phone             & guitar            & tool              & art               \\ \hline
No-Fed-Rand-Init-GNN     & 0.687 $\pm$ 0.005 & 0.709 $\pm$ 0.008 & 0.673 $\pm$ 0.005 & 0.737 $\pm$ 0.014 & 0.622 $\pm$ 0.008 & 0.653 $\pm$ 0.009 \\
No-Fed-Sup               & 0.787 $\pm$ 0.009 & 0.801 $\pm$ 0.008 & 0.759 $\pm$ 0.007 & 0.866 $\pm$ 0.012 & 0.764 $\pm$ 0.012 & 0.741 $\pm$ 0.010 \\
No-Fed-Self-Freeze       & 0.837 $\pm$ 0.004 & 0.841 $\pm$ 0.007 & \textbf{0.778 $\pm$ 0.003} & 0.882 $\pm$ 0.011 & 0.800 $\pm$ 0.007 & 0.784 $\pm$ 0.006 \\
No-Fed-Self-Finetune     & 0.789 $\pm$ 0.012 & 0.792 $\pm$ 0.014 & 0.757 $\pm$ 0.012 & 0.845 $\pm$ 0.026 & 0.758 $\pm$ 0.018 & 0.735 $\pm$ 0.015 \\ \hline
Fed-Sup                  & 0.769 $\pm$ 0.003 & 0.754 $\pm$ 0.009 & 0.741 $\pm$ 0.004 & 0.807 $\pm$ 0.019 & 0.722 $\pm$ 0.010 & 0.704 $\pm$ 0.013 \\
\textbf{Fed-Self-Freeze} (ours)   & \textbf{0.849 $\pm$ 0.005} & \textbf{0.868 $\pm$ 0.009} & 0.776 $\pm$ 0.005 & \textbf{0.897 $\pm$ 0.011} & \textbf{0.818 $\pm$ 0.010} & \textbf{0.807 $\pm$ 0.005} \\
Fed-Self-Finetune (ours) & 0.786 $\pm$ 0.009 & 0.791 $\pm$ 0.012 & 0.753 $\pm$ 0.005 & 0.849 $\pm$ 0.019 & 0.764 $\pm$ 0.016 & 0.738 $\pm$ 0.015 \\ \hline
\end{tabular}%
}
\end{table*}

\begin{figure}
    \centering
    \includegraphics[width=0.5\textwidth]{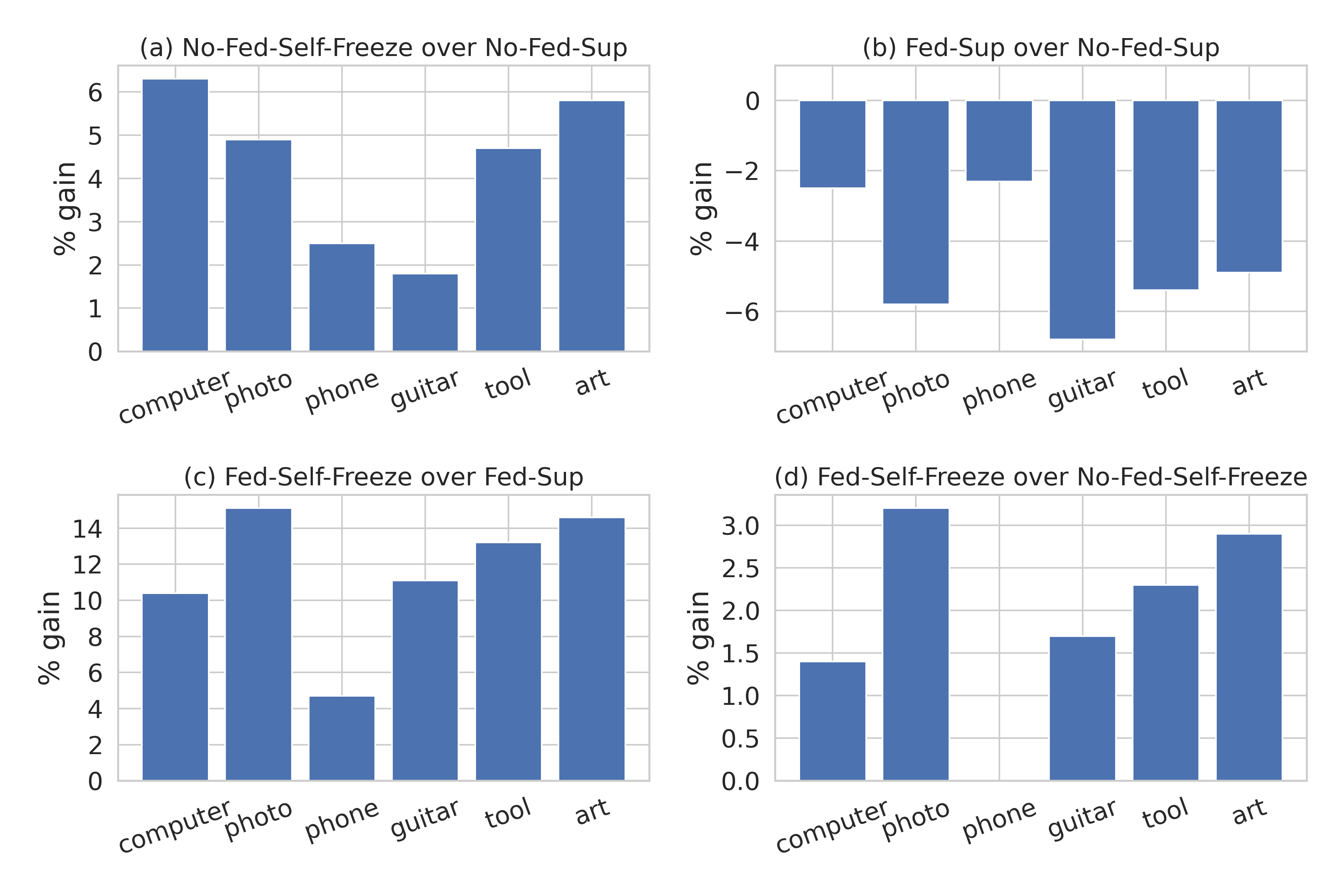}
    \caption{Performance gains in Amazon network experiments}
    \label{fig:gains_amazon}
\end{figure}

Table~\ref{tab:amzn_results} shows the results across all Amazon client graphs. Firstly,  No-Fed-Self-Freeze is clearly superior to No-Fed-Sup, which shows that self-supervised learning is effective for these graphs with an avg., gain of 4.3\% across all clients and  Fig~\ref{fig:gains_amazon}(a) shows the gains individually. Note that, the number of train node labels for amazon graphs are extremely limited---we consider only 10\% of all nodes as train nodes. In this case, No-Fed-Sup suffers from over-fitting, where as No-Fed-Self is able to utilize the unlabeled nodes effectively---thereby justifying the SSL objective. Secondly, No-Fed-Self-Finetune is worse than No-Fed-Self-Freeze, indicating that fine-tuning is not very effective for Amazon networks which is contrary to what we observed in Twitch. We attribute this again to limited train labels and fine-tuning is still ineffective. Thirdly, we interestingly find that Fed-Sup suffers very badly across all clients and we witness avg. gain of -4.16\% compared to No-Fed-Sup across clients. Fig~\ref{fig:gains_amazon}(b) further shows it individually. This is again different to what we observed in twitch where the same comparison (Fed-Sup vs No-Fed-Sup) did not lead to this kind of severe performance degradation. We reason that because of the increased downstream task heterogeneity in Amazon compared to Twitch, a global model learnt using label supervision alone will be incapable to perform well. Moreover, The non-IID ness among Amazon clients is significantly greater than among Twitch clients and standard supervised FedAvg (Fed-Sup) falls short. Fourthly, our method \textbf{Fed-Self-Freeze} provides significant increase in performance compared to baselines across all clients due to the effective and collaborative use of unlabeled data. Specifically, we witness avg. 11.5\% gains across clients over Fed-Sup which shows how our formulation can be effectively used when there is high task heterogeneity across clients. Fig~\ref{fig:gains_amazon}(c) further shows the gains individually. We also witness an avg. gain of 1.9\% over No-Fed-Self-Freeze and further shows that our formulation can indeed leverage unlabeled data across clients using federation to improve downstream performance locally.        

\section{Conclusion}
Motivated by two key characteristics of real-world graph data when used in cross-silo settings: (1) Label deficiency and (2) Downstream task heterogeneity, we proposed a novel problem setting that has not been considered by current FedGRL setups. We raise a natural research question of how one can leverage vast amounts of unlabeled graph data and collaboratively learn a shared model using federated learning that is effective at solving downstream client tasks. We provide a general formulation based on model interpolation where a shared global model is both self-supervised and gets supervision with available local task labels through a local client model. We provided two learning protocols based on this general formulation and provided specific instantiations using BGRL for the self-supervised objective. To empirically verify one of our learning protocol and its instantiation, we adopted a real world Twitch Gamer Networks to simulate a cross-geo FedGRL application and we witnessed on avg. 6.1\% gains over traditional supervised federated objectives. To further incorporate the task heterogeneity characteristic, we constructed a new cross-silo dataset called Amazon Co-purchase Networks which have varying downstream label domains. We showed how standard supervised federated objectives can result in negative gains (on avg. -4.16\%) compared to individual client specific training, due to the increased data heterogeneity and finally justified our formulation which results in on avg. 11.5\% gains over traditional supervised federated learning and on avg. 1.9\% gains over individually trained self-supervised models. 

Several extensions to our work can be considered. While we have experimented with learning protocol \textbf{LP1} in this paper, more experiments are required to fully realize the potential benefits of the formulations, especially \textbf{LP2}. It is also significantly more challenging as the shared global model may be biased towards local client labels due to supervision and techniques from personalized federated learning literature may have to be adapted, instead of the standard FedAvg algorithm we currently employ. Further work can also be undertaken to explore the best server aggregation protocol, experiments with varying client participation ratios and how it affects self-supervised based federation can also be further explored. Moreover, our methodology of constructing the amazon co-purchase networks can be extended to include more clients for future bench-marking. We stress on the fact that for FedGRL, realistic datasets are a major gap in current works and more effort is required in this front.

\bibliographystyle{plain}
\bibliography{base}


\end{document}